%% file: paper.tex
\documentclass{article}
\usepackage{PRIMEarxiv}

\usepackage{authblk}
\usepackage[utf8]{inputenc} % allow utf-8 input
\usepackage[T1]{fontenc}    % use 8-bit T1 fonts
\usepackage{hyperref}       % hyperlinks
\usepackage{url}            % simple URL typesetting
\usepackage{booktabs}       % professional-quality tables
\usepackage{amsfonts}       % blackboard math symbols
\usepackage{nicefrac}       % compact symbols for 1/2, etc.
\usepackage{microtype}      % microtypography
\usepackage{lipsum}
\usepackage{fancyhdr}       % header
\usepackage{graphicx}       % graphics
\graphicspath{{media/}}     % or\usepackage{graphicx}
\usepackage{xspace}
\usepackage{stackengine}
\usepackage{multicol}
\usepackage{bm}
\usepackage{siunitx}

\usepackage{multirow}
\usepackage{breakcites}
\usepackage{float}
\usepackage{amsmath}
\usepackage{mathtools} % amsmath with fixes and additions
\usepackage{booktabs} % commands to create good-looking tables
\usepackage{tikz} % nice language for creating drawings and diagrams
\usepackage{graphicx} % DO NOT CHANGE THIS
\usepackage{xcolor}
\usepackage{hyperref}
\usepackage{amsthm}
\usepackage{array}
\newcolumntype{P}[1]{>{\centering\arraybackslash}p{#1}}

\usepackage[toc,page]{appendix}

\usepackage[utf8]{inputenc} % 
\usepackage{url}            % simple URL typesetting
\usepackage{booktabs}       % professional-quality tables
\usepackage{amsfonts}       % blackboard math symbols
\usepackage{nicefrac}       % compact symbols for 1/2, etc.
\usepackage{microtype}      % microtypography
\usepackage{xcolor}         % colors
\usepackage{multirow}
\usepackage{makecell}
\usepackage{cleveref}
\usepackage[caption=false]{subfig}

\title{Beyond Eviction Prediction: Leveraging Local Spatiotemporal Public Records to Inform Action}

\author[1]{Tasfia Mashiat}
\author[2]{Alex DiChristofano}
\author[2]{Patrick J. Fowler}
\author[1]{Sanmay Das}
\affil[1]{George Mason University. Email:\{tmashiat, sanmay\}@gmu.edu}
\affil[2]{Washington University in St. Louis. Email:\{a.dichristofano, pjfowler\}@wustl.edu}

\begin{document}

\maketitle

\begin{abstract}
There has been considerable recent interest in scoring properties on the basis of eviction risk. The success of methods for eviction prediction is typically evaluated using different measures of predictive accuracy. However, the underlying goal of such prediction is to direct appropriate assistance to households that may be at greater risk so they remain stably housed. Thus, we must ask the question of how useful such predictions are in targeting outreach efforts -- informing action. In this paper, we investigate this question using a novel dataset that matches information on properties, evictions, and owners. We perform an eviction prediction task to produce risk scores and then use these risk scores to plan targeted outreach policies. We show that the risk scores are, in fact, useful, enabling a theoretical team of caseworkers to reach more eviction-prone properties in the same amount of time, compared to outreach policies that are either neighborhood-based or focus on buildings with a recent history of evictions. We also discuss the importance of neighborhood and ownership features in both risk prediction and targeted outreach.
\end{abstract}


\input{sections/1-Introduction}

\input{sections/2-StudyContext-Data}

\input{sections/3-Prediction}

\input{sections/4-Outreach}

\input{sections/5-Discussion}

\input{sections/6-Statements}

\section*{Acknowledgments}
The St. Louis Regional Response Team and members of the collaborative network offer ongoing assistance in the research design. We are grateful for support from NSF grants 2127752 and 1939677, as well as support from Amazon through an NSF/Amazon FAI Award.

\bibliographystyle{unsrt}

\bibliography{references}

\input{sections/7-Appendix}

\end{document}

%% file: sections/1-Introduction.tex
\section{Introduction}

One of the standard methodologies for using AI in the social service provision, education, and criminal justice domains is for algorithms to generate risk scores used to direct prioritization metrics or as additional information in bureaucratic decision-making procedures. Examples include the investigation of possible child maltreatment or neglect \cite{chouldechova2018case}, homelessness resource provision \cite{kube2019allocating,azizi2018designing}, sentencing decisions \cite{pruss2023ghosting}, and targeting interventions to students at risk of dropping out of school \cite{perdomo2023difficult}. 

A pressing unresolved question is how these possible applications of AI should be evaluated. First, of course, it is important to compare with the bureaucratic counterfactual \cite{Johnson_Zhang_2022, Pokharel_Das_Fowler_2023} as well as to understand the impact of how information is presented to and used by the human-in-the-loop \cite{kube2022just,narayanan2023does}. Even beyond this, though, there is a question of whether improvements in the oft-studied accuracy of the predictive model \cite{tabar2022forecasting, ye2019using} are truly useful in decision-making \cite{Wilder_Dilkina_Tambe_2019}. For example, recent work in the context of education suggests that even if a model accurately sorts students in terms of their dropout risk, it may provide no benefit in targeting interventions over a model that only uses information about students' environments \cite{perdomo2023difficult}. However, the contextual and institutional specifics may be important. Our goal in this paper is to study the problem of targeting interventions based on risk scoring in the context of tenant eviction. 

We aim to assess and inform data-driven targeting of assistance to renters at greater risk for eviction. The goal is to keep these renters stably housed. %through court proceedings with the goal of maintaining stable housing. 
Currently, most communities target tenant outreach at the neighborhood level based on historical eviction rates (where available) or geographic characteristics (e.g., zip code poverty levels). Few communities use dynamic spatiotemporal information or predictive modeling for identifying at-risk properties. Prior research on eviction prediction has used somewhat imprecise data 
(e.g., zip code versus property-level prediction, lack of property ownership profiles)
and has been measured only in terms of predictive accuracy rather than effectiveness in designing outreach policies~\cite{tabar2022forecasting}. 

In this work, we leverage a unique community-based research collaboration to build predictive models of eviction and then use these to design and analyze potential outreach policies.
Our results also help us better understand the importance of different types of features in prediction and outreach design to keep tenants housed.

\noindent \textbf{Contributions: }
We construct a novel dataset, combining extensive spatiotemporal data that no prior works possess. Our community partners assembled available information sources to inform decision-making on targeting health and economic support to households most adversely impacted by the pandemic. Our partners have aggregated public and proprietary historical information on evictions (e.g., parties, dates of filing and judgment, etc.) and building information (e.g., units, owner, etc.) across all properties at St. Louis City and County in Missouri, USA. Matching owner addresses from tax assessments allows us to create unique owner profiles (e.g., business status, owner location, number of units, etc.) unavailable in most communities. The geocoded data can also easily integrate relevant neighborhood characteristics at the block, block group, and tract levels available through the Census and other sources, such as racial segregation, poverty, unemployment, etc. 

Using machine learning methods, we leverage the assembled spatiotemporal records to predict court eviction filings at the property level. Modeling employs three machine learning classifiers (i.e., Random Forest, XGBoost, and Feedforward Neural Networks) and iteratively incorporates 1) historical eviction records, 2) neighborhood features, and 3) owner characteristics to assess improvements in predictive performance with additional features. We assess accuracy over time by comparing the change in predictive performance at 3, 12, and 24 months following training.

Finally, we turn to the evaluation of the effectiveness of risk scores produced by the models in targeting outreach efforts -- identifying households with a high risk of eviction and routing a theoretical set of caseworkers to the households for assistance. 
We compare risk-score-based outreach policies with commonly used alternatives: outreach based on the recent history of evictions in 1) individual households and 2) neighborhoods.

\textbf{Preview of Results:} Our findings show high predictive accuracy for evictions in St. Louis.
Predictive accuracy improves when we incrementally add owner and neighborhood information as features. For example, using only historical eviction records, XGBoost yields an area under the ROC curve of $0.76$. Incorporating the neighborhood features with the eviction records improves this by $16\%$ (to $0.84$). The AUC further increases by $11\%$ (to $0.89$) when we integrate owner features into the model. We show this trend remains relatively stable at 3, 12, and 24 months after training when using all information, whereas performance degrades rapidly when using historical eviction records alone. Models generalize well even when the COVID shock is part of the intervening time frame.

More importantly, we are able to analyze the use case -- targeting outreach using our risk scores. When simulating a targeted outreach policy that focuses on higher-risk households, we show that caseworkers would reach $8.5\%$ more eviction-prone properties in the same amount of time as compared with neighborhood-based outreach and $28\%$ more than an outreach policy focused on buildings with a history of recent evictions. The addition of neighborhood features (demographics, socioeconomic status, and characteristics of the housing stock) adds substantial marginal value to the risk scores when used in our routing policy, while the further addition of owner attributes has limited additional value beyond that. We are also able to report that the improvement in outreach policy performance shows a much stronger correlation with improvement in the area under the ROC curve, as opposed to improvement in the area under the precision-recall curve, despite literature claiming benefits for the latter in low base-rate tasks such as ours \cite{davis2006relationship}. Importantly, given the context, our results support the feasibility of using risk modeling and data-driven targeting to inform eviction prevention. 

\subsection{Social Significance and Prior Research} 

The lack of affordable housing threatens the safety and security of low-income households across the US. Before COVID-19, an estimated 36.5 million US households experienced housing burden, defined as paying more than 30\% of income on rent or mortgages \cite{joint_center2020}. The Census Household Pulse Survey -- a nationally representative biweekly assessment of the economic and health impacts of COVID-19 -- shows that 30\% of all renters felt little or no confidence in paying the next month's rent three years into the pandemic, while racial disparities in housing insecurity persist \cite{kim2021financial, pulse2021}.

In addition to its obvious economic implications, eviction has significant negative effects on health and well-being, including higher rates of stress, depression, anxiety, and substance abuse with life-course implications not just for adults but also for the health and well-being of children \cite{desmond2016evicted, himmelstein2021eviction, tsai2019systematic}. Several studies using the Pulse Survey show that households at risk for housing displacement report elevated rates of anxiety and depression \cite{acharya2022risk, bushman2022housing, kim2021financial}.
Concerns regarding a tidal wave of evictions at the onset of the COVID-19 pandemic triggered unprecedented federal responses, including federal, state, and municipal moratoria on evictions and allocation of \$46 billion of housing assistance for low-income households. The policies implemented in this response, including moratoria on eviction filings and executions and delivery of rental assistance, improved outcomes \cite{boen2023buffering, donnelly2021state, leifheit2021variation}. 

Today, however, eviction filing rates continue to rise beyond pre-COVID levels in communities across the country -- putting 2.3 million renters at risk in the near future \cite{gromis2022estimating, marccal2023feedback}. Governmental and non-governmental agencies continue to actively search for ways to protect renters from the devastating impact of eviction on health and well-being \cite{desmond2015eviction,bieretz2020getting,desmond2022unaffordable,desmond2016evicted}.

Community efforts to prevent evictions typically focus on providing tenants with time-limited assistance, such as one-time financial aid to pay rental and utility arrears, landlord mediation to resolve disputes, legal representation in court proceedings, or referral to an array of local services provided by governmental and non-governmental agencies \cite{collins2022we, holl2016interventions, woellenstein2022homelessness}. The lack of affordable housing in many communities challenges the efficient targeting of resources, as demand for stable housing outpaces the availability of limited resources \cite{fowler2019solving, shinn2020midst}. Governmental and non-governmental agencies struggle to coordinate in identifying and responding to housing risk that warrants intervention with available assistance. Machine learning techniques applied to eviction prediction offer promise for improving the efficiency of prevention efforts; however, technical and ethical challenges impede efforts. Data on rental market dynamics generally lack specificity to inform targeting responses. For example, few municipalities track which residential housing includes rental units, and court eviction records -- where available -- capture a fraction of places where evictions may occur, impeding community-wide surveillance. The most proximal information on households collected through the Census, such as race, socioeconomic status (SES), building features, etc., is captured at the block level, or a statistical area of hundreds of housing units, and averaged over five years. Proprietary records on rental units (e.g., average rent listings) collected by Zillow and other sources are reliable at the zip code level -- limiting the application of real-time responses to market shifts. Even municipal records collected through tax assessments, code enforcement, and other governmental interactions fail to identify the properties held by a single owner registered under various names and limited liability corporations (LLCs), which obscures shared risk across properties and potential landlord interventions. 

Although limited in application, initial efforts to use algorithmic approaches for risk detection show promise. One study in Dallas, Texas, predicted eviction rates within census tracts, representing statistical areas including 4,000 residents on average~\cite{tabar2022forecasting}. A neural network built with historical court filing records and an array of tract-level sociodemographic characteristics produced reasonable predictive accuracy that generalized to unseen census tracts. Yet, the tract-level results provide limited practical utility for targeting households for prevention. Another study conducted before COVID-19 in New York City captured building-level risk of landlord harassment reported by tenants to the Mayor's Office \cite{ye2019using}. A gradient-boosting classifier trained on 92 building-level and neighborhood characteristics outperformed other machine-learning classifiers and expert-driven canvassing in predicting the likelihood of harassment. The results demonstrate the utility of data-driven approaches despite not directly targeting risk for eviction.

Several barriers impede the broader implementation of risk detection and data-driven eviction prevention. Assembling public and proprietary information for model building requires extensive integration of disparate data systems designed for different purposes. For instance, court records of evictions document procedural processes for adjudication without the intention of linking to other municipal data, such as tax assessment and code enforcement. Beyond the nontrivial effort of data collection, much remains uncertain in building accurate models. The likelihood of eviction depends on multiple interacting contexts that prove challenging to capture \cite{tabar2022forecasting, ye2019using}.  
Moreover, little is known regarding whether and how rental property owner characteristics (e.g., number of properties, corporate status, location) matter for evictions; sociological research on predatory behaviors of slumlords would suggest so \cite{desmond2016evicted}. Likewise, the accuracy of eviction risk predictions over time remains unclear. Pragmatically, it would be easier to deploy interventions based on models that require infrequent updating; however, rapid rental market dynamics could degrade prediction accuracy quickly, requiring rapid adjustments. The fallout from COVID and slow economic recovery warrants careful attention in building accurate models. The present study addresses these questions through a novel community-based research partnership.

%% file: sections/2-StudyContext-Data.tex
\section{Study Context, Design, and Data}
\label{sec:study_context_data}

\subsection{Study Context}
The study leverages an ongoing community-based research partnership aimed at preventing tenant evictions in St. Louis, MO. In combination with the neighboring (and non-overlapping) St. Louis County, the midsize, Midwestern city includes 1.3 million residents -- 61\% of whom identify as White and 30\% as Black. Racial and economic segregation remains challenging in St. Louis. Approximately 30\% of the population lives in a neighborhood that is more than 90\% White or Black, and Black households are nearly three times more likely to fall below the federal poverty level. Among the approximately 210,000 renting households in St. Louis, 
 46.5\% and 47\% are rent-burdened in the City and County respectively, meaning that they spend 30\% or more of their income on rent~\cite{censusdata}. Community partners include academics, housing and social service providers, governmental and non-governmental administrators, local philanthropic agents, and a civic technology firm with extensive expertise in linking and spatially mapping property-level information collected across municipal departments.

\begin{figure}
     \centering
     \includegraphics[width=0.65\columnwidth]{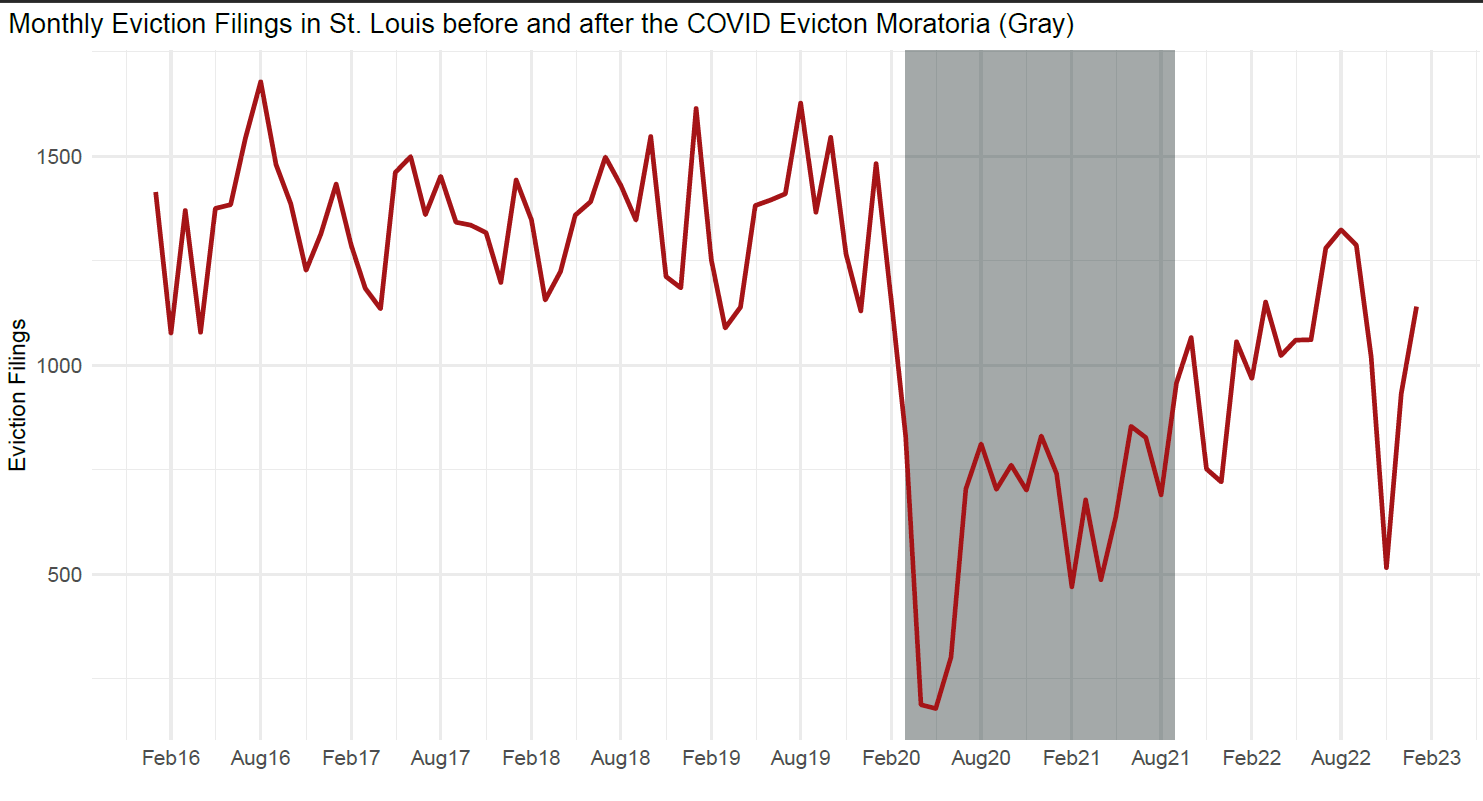} 
     \caption{Monthly eviction filings in St. Louis from January 2016 - January 2023.}
    \label{fig:eviction_trend}
\end{figure}

\Cref{fig:eviction_trend} illustrates eviction filing trends from January 2016 through January 2023. Before COVID-19, filings averaged eight evictions per year per 100 renters. Eviction moratoria between March 2020 and August 2021 prohibited the execution of evictions and slowed filing rates that gradually returned to pre-COVID levels by the end of 2023. 

\subsection{Study Design}
We present a two-stage empirical study. First, we evaluate property-level eviction prediction performance for different predictive models; specifically, we build binary classifiers of whether an eviction occurs at a residential property. Properties are defined as the smallest separable taxable unit of real estate. In a broader setting, properties take the form of individual homes, entire apartment buildings, or single units of a condominium or townhome. In this work, we restrict the study to properties with two or more units, given differences in the frequencies of evictions; filings occur at less than one-quarter the rate within single-unit properties, according to court records.

\begin{figure}[t]
     \centering
     \includegraphics[width=0.5\columnwidth]{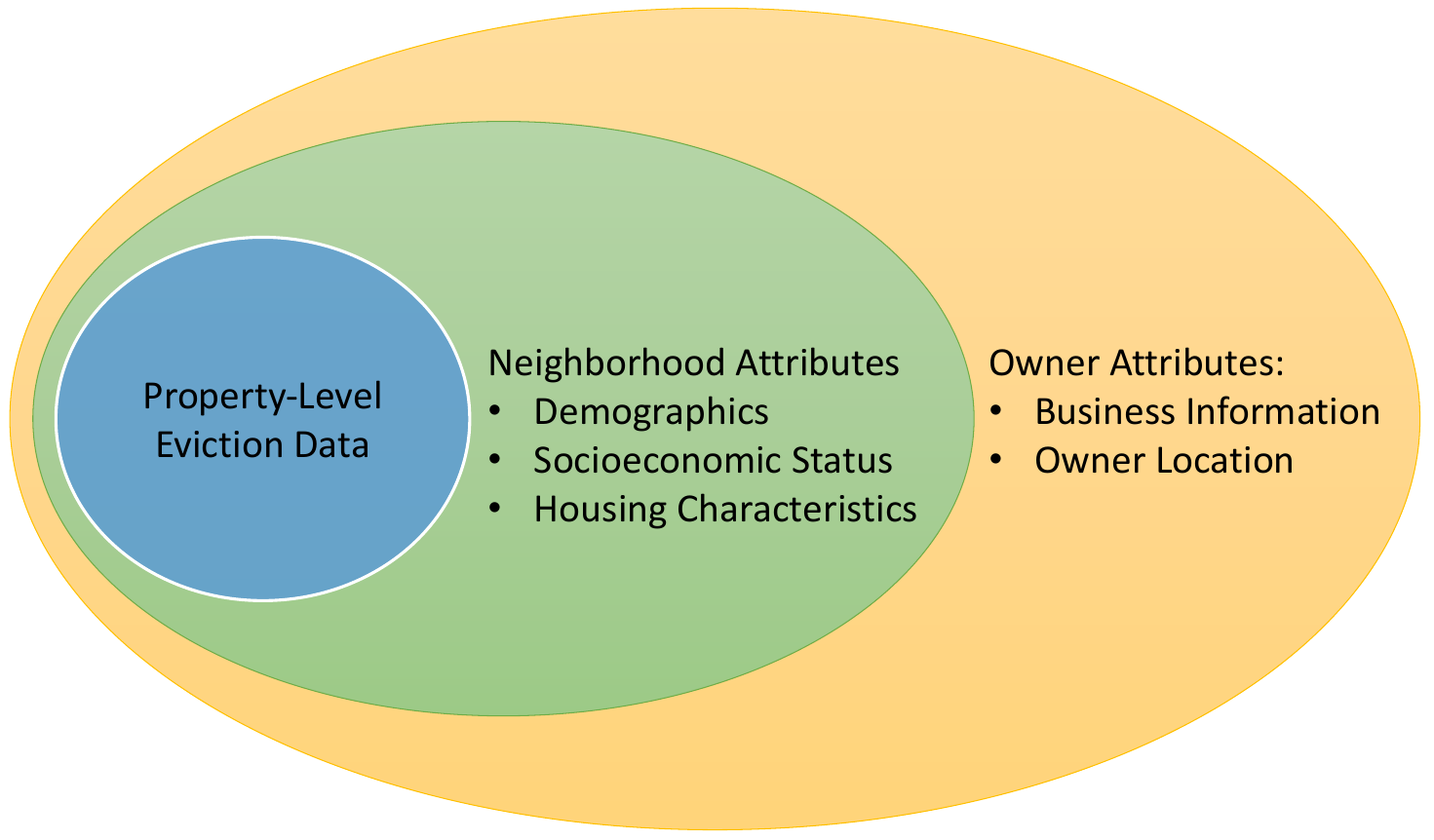}  
    \caption{Thematic representation of the feature sets in different models. 
      Each property is associated with an eviction record and is situated within a neighborhood, specifically a Census block and block group. Properties are owned by owners who may own multiple properties in different neighborhoods. The number of features incorporated into the binary classifier expands with the radius of the diagram.}
    \label{fig:thematic_representation} 
\end{figure}

\Cref{fig:thematic_representation} illustrates the three feature categories of focus -- the historical eviction patterns of the property, neighborhood attributes such as demographics and income, and owner information. We analyze whether incrementally adding these features to eviction records affects prediction.  
We also evaluate how model performance changes with both increased time between training and testing periods and pandemic-induced shifts in the local housing market, which include eviction moratoria and allocation of emergency rental assistance.

In the second stage, we utilize risk scores generated from the best-performing model to design and test outreach routing policies that target vulnerable tenants before an eviction filing occurs. Simulations compare our risk-score-based targeted outreach policy with commonly employed outreach practices.
We evaluate the policies based on the number of evictions discovered in a fixed period.

\subsection{Data}

\input{tables/table-1}

Our civic technology partner assembled a map of all properties in the region and joined these properties with information from multiple sources using standard name and definition protocols. Data came from governmental and nongovernmental open data portals that periodically pulled eviction records (daily), tax assessments (annually), and other information when made available. Data were spatially linked through GIS and merged with Census features. Extensive verification identified owners by linking names and mailing addresses in tax assessments available from 2019 through 2023. Each owner received an owner identification number assigned across all regional properties. Ownership changes were captured temporally. 

We combine these data\footnote{Part of our data is proprietary and subject to a specific data sharing agreement that does not allow for public release.} to predict eviction at a property. \Cref{Tab1:Feature_description} describes the attributes used.

\paragraph{Eviction Filings} Court-involved eviction filings represent the primary outcome. By overlaying eviction records across the entire geographic area, we leverage address-level data to capture evictions within properties that can be linked with building, neighborhood, and owner characteristics.

There were 45,197 mappable eviction filings across 15,349 city and county properties from January 2019 through February 2023\footnote{An additional 4,351 filings (8\%) could not be mapped from court records.}. Each eviction record includes a filing date, case ID, property ID, and information on the defendant, plaintiff, and plaintiff's attorney. 
As detailed below, we counted the monthly evictions for each property at distinct periods and generated the target labels of whether an eviction occurred in subsequent periods.

\paragraph{Neighborhood Characteristics} We link each property with five-year block group estimates collected through the 2019 American Community Survey (ACS)\footnote{https://www.census.gov/programs-surveys/acs/data.html}. Block groups -- the smallest geographic units released by the Census for the ACS -- typically represent 600 to 3,000 residents and capture neighborhood features, such as median income, gross rent, households receiving food stamps or cash assistance, and health insurance. We use more proximal block-level attributes for race/ethnicity, age, and occupancy status available through the 2020 Decennial Census\footnote{https://www.census.gov/programs-surveys/decennial-census/decade/2020/2020-census-main.html}. 
 Rates were computed to capture neighborhood characteristics at the lowest level of geography, such as the percentages of each race within a block (Decennial Census) and high school graduates in a block group (ACS). Table~\ref{Tab1:Feature_description} lists all neighborhood attributes included in modeling.

\paragraph{Owner Characteristics} We selected properties listed as residential in tax assessments with more than two units within the property. There are 26,770 properties that meet the criteria with linked owner information.
Tax assessments record the initiation and termination dates for property ownership, allowing linkages between a property and the owner in a given time frame. Each owner-property pair includes records on whether the owner is a business, the business type (e.g., LLC versus estate), if the owner lives out-of-state, and whether the owners reside at the property. Moreover, we identified whether owners worked with high eviction filing attorneys. Court records enabled counting filings by each attorney, which we ranked from highest to lowest for a given period. Two time-varying indicators captured whether owners filed at least one eviction with (1) a top 25 highest-filing attorney or (2) a top 26 to 50 highest-filing attorney in a given period.

%% file: tables/table-1.tex
\begin{table*}
\centering
\small%
\def\arraystretch{1.2}%
\caption{Attribute Description}

\begin{center}
  \begin{tabular}{|c|p{10 cm }|c|}
  \hline
  \centering\textbf{Category} & \centering\textbf{Attribute Details} & \textbf{Data Type}\\
  \hline
  {Eviction Filings} &
  \# of evictions during the period by month and quarter & Integer\\

  \hline
   & \textbf{Decennial Census Block Data:} \% of population under 18; \% of housing units occupied; \% of population by racial and ethnic groups  & \\

  \thead[c]{Neighborhood \\Characteristics}& \textbf{American Community Survey Census Block Group Data:} median household income; median gross rent; gross rent as a \% of household income (GRAPI); \% of housing units that are renter occupied; \% of renter-occupied housing units with more than one occupant; \% of adults below the federal poverty level; \% of households with a mortgage; \% of households receiving SNAP and/or public assistance; \% of households with health insurance; \% of female-headed households with children under 18; \% of adults with a high school degree; \% of adults who are a veteran & Float\\
  \hline

   & \# of units in the property & Integer \\
  \cline{2-3}
  & \# of properties in St. Louis held by the owner  & Integer\\
  \cline{2-3}
 \multirow{3}{*}{\thead[c]{Owner \\Characteristics}} & is the owner a business  & Boolean\\
 
  \cline{2-3}
  & is the property owner-occupied  & Boolean\\
  \cline{2-3}
  & has the owner used one of the top $25$ highest-filing attorneys during the period & Boolean\\
\cline{2-3}
  & has the owner used one of the top $50$ highest-filing attorneys during the period & Boolean\\

  \cline{2-3}
  & does the owner live in St. Louis, in-state, or out of state  & Categorical\\
  \hline
  
  \end{tabular}
  \end{center}
  \label{Tab1:Feature_description}
\end{table*}

%% file: sections/3-Prediction.tex
\section{Predicting Risk Scores for Eviction}

\begin{figure}[!htb]
     \centering
     \subfloat[Training and Test periods for Pre-COVID Model]{\includegraphics[width=0.56\columnwidth]{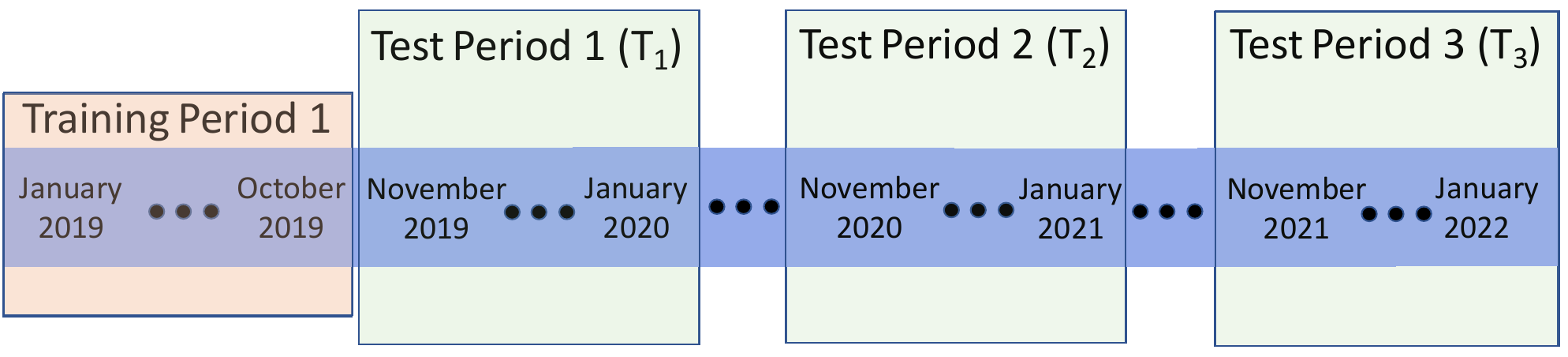}\label{fig:pre_covid_timeframe} }  
 
     \subfloat[Training and Test periods for Post-COVID Model]{\includegraphics[width=0.47\columnwidth]{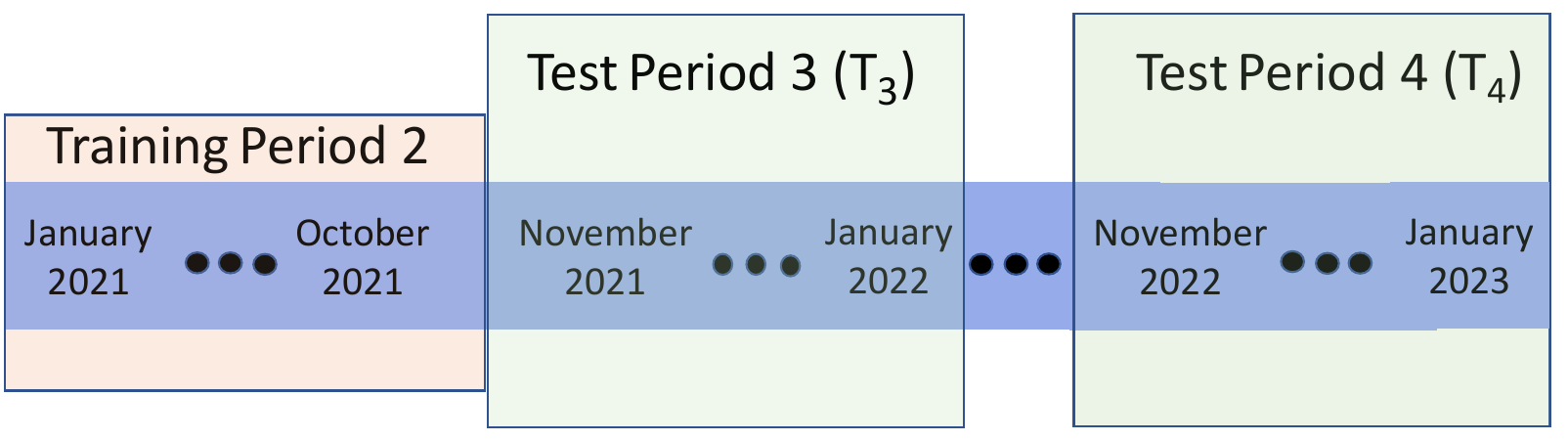}\label{fig:post_covid_timeframe} }
  
    \caption{Training and testing timelines. The three black dots represent the interim months between the beginning and end of a period. There are two training periods and four testing periods. We do not train a model on data from 2020 due to the outsized influence of COVID-19 and the corresponding eviction moratoria on eviction patterns. We denote the model trained using Training Period 1 as the \textbf{Pre-COVID} Model and the model trained using Training Period 2 as the \textbf{Post-COVID} Model. We select three-month test periods composed of November, December, and January to compare the ability of the two models to generalize into the future. Data from all testing periods are withheld from both models during training.}
    \label{fig:timeframes}
\end{figure}

\input{tables/table-2}
\subsection{Experimental Setup}

\paragraph{Models} 
The eviction models built in this study generate property-level scores valued between 0 and 1 representing the risk of an eviction occurring within the next three months. Eighteen classification models were built, each leveraging one of three different learning algorithms, incorporating one of three nested subsets of features, and trained on data from either before or after the start of the COVID-19 pandemic. The three algorithms used were random forests (RF), gradient-boosted decision trees via XGBoost (XBG), and a Feedforward Neural Network (FNN). Implementation details for these models are described in Section~\ref{sec:model_details} of the Appendix. The sets of features are visualized in \Cref{fig:thematic_representation} and described in detail in \Cref{Tab1:Feature_description} (Eviction Filings; Eviction Filings and Neighborhood Characteristics; Eviction Filings, Neighborhood Characteristics, and Owner Characteristics), while the model timeline is discussed in \Cref{sec:model_timeline}.

\paragraph{Train-Test Splits and Model Timeline} \label{sec:model_timeline}
\Cref{fig:timeframes} illustrates the model timeline with training and test set splits. We drew data to train the models from two ten-month periods: \textbf{Pre-COVID}, the time from January 2019 to October 2019, and \textbf{Post-COVID}, the time between January 2021 to October 2021. During training, models predict whether an eviction occurred in the last three months of the training period (August-October) by utilizing features from the first seven months of the period (January-July).
Feature sets (i.e., historical eviction filings, neighborhood characteristics, and owner characteristics) are included iteratively with five-fold cross-validation.

We test our models using three-month intervals following training. The \textbf{Pre-COVID} models test at three months from November 2019 to January 2020 ($T_{1}$), twelve months from November 2020 to January 2021 ($T_{2}$), and 24 months from November 2021 to January 2022 ($T_{3}$). \textbf{Post-COVID} models test three months ($T_{3}$) following training and twelve months from November 2022 to January 2023 ($T_{4}$).

\paragraph{Evaluation Metrics} 
We first evaluate model performance by plotting the Receiver Operating Characteristic (ROC) curves. The ROC Area Under the Curve (AUC) provides a base-rate agnostic measure for differentiating between positive (Eviction) and negative cases; AUC values reaching 1 indicate perfect differentiation between the two cases in the sense that all the positive examples are ranked higher than the negative examples.

We also assess the AUC of the PR curve, which reports the proportion of positive cases (evictions) correctly identified for different probability thresholds \cite{cook2020consult}. Precision measures how many of the items predicted to be positive cases are truly positive cases (true evictions), and recall is the fraction of positive cases identified. 
As the base rates of evictions are low $(<0.1)$, %(\Cref{fig:comparative_results}), 
the area under the PR curve could provide a useful measure for informing the value of interventions when compared against the base rate \cite{davis2006relationship,sofaer2019area}. We later investigate further the value of both metrics in relation to the actual outreach goal. 
We use DeLong’s algorithm to test the statistical significance of differences in ROC and PR AUCs between models \cite{delong1988comparing}.

\begin{figure*}
    \centering
    \captionsetup[subfigure]{justification=centering}
    \centering
    
    \subfloat[ROC Curve for Pre-COVID (XGB) Model \\ and $T_{1}$ Prediction]{\includegraphics[width=.4\textwidth]{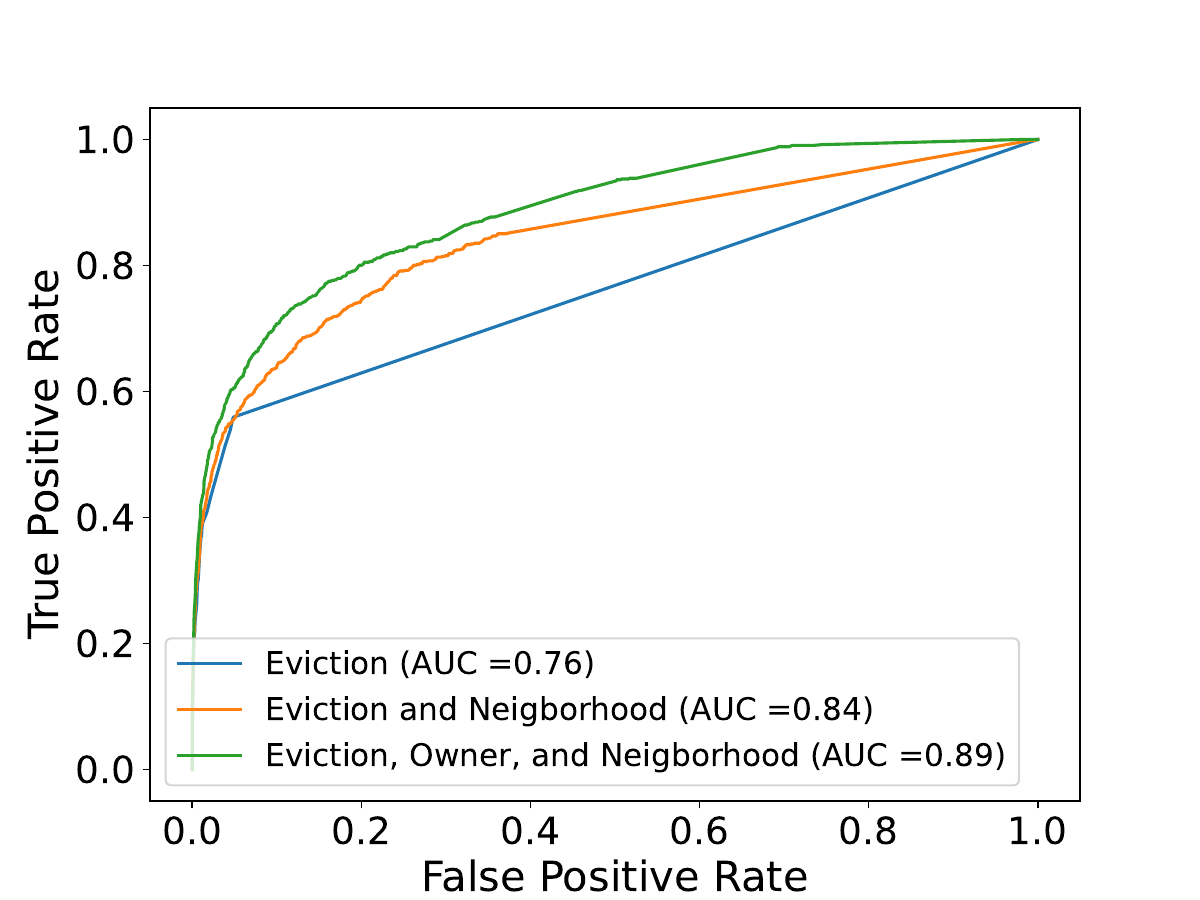}\label{fig:xgb_roc_19}}
    \subfloat[ROC Curve for Post-COVID (XGB) Model \\ and $T_{3}$ Prediction]{\includegraphics[width=.4\textwidth]{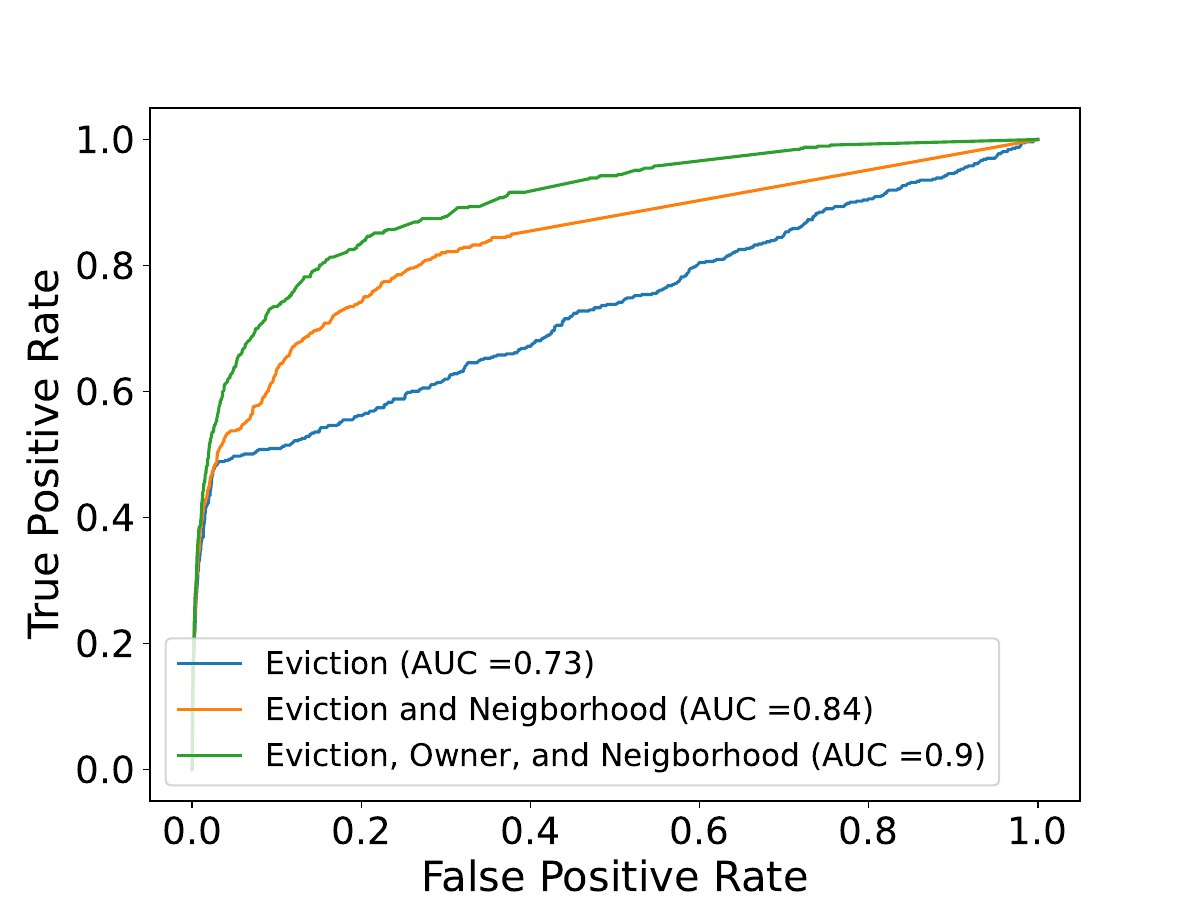}\label{fig:xgb_roc_21}}
    \caption{ROC Curves for $T_{1}$ and $T_{3}$. (a) The AUC value for $T_{1}$ prediction using the Pre-COVID (XGB) model that includes Eviction, Owner, and Neighborhood attributes has the highest ROC AUC ($0.89$), and outperforms the baseline and model trained on eviction+neighborhood with high levels of statistical significance (p-values of $3.34e^{-72}$ and $2.44e^{-31}$ respectively); (b) Inclusion of Owner, and Neighborhood attributes in the Post-COVID (XGB) model also increases the ROC AUC and the curve is significantly different than the baseline and the model trained on including neighborhood attributes with p-values $1.27e^{-74}$ and $8.00e^{-19}$.}
    
     \label{fig:roc_comparison}
\end{figure*}

\subsection{Prediction Results and Analysis}
Table~\ref{tab:comparative_results} compares the predictive performance of the risk models. We analyze performance on different feature sets, model structures (decision trees vs. FNN), timeframes, and evaluation metrics.

\textbf{Neighborhood and Owner Features Improve Performance.} We start by comparing the models with different feature sets in terms of their performance in predicting the immediate future (using features from the last seven months to predict if there will be
an eviction in the next three). For the Pre-COVID (XGB) model, when using only prior eviction records without any information about the owner or the property, the ROC AUC for $T_{1}$ is $0.76$. Adding the neighborhood attributes yields a $8$-point increase in the AUC score, while including the owner attributes as features achieves an additional $5$-point gain in AUC. Figure~\ref{fig:xgb_roc_19} shows the ROC curves for all three cases. A similar pattern emerges for the Pre-COVID (RF) and Pre-COVID (FNN) models. For both models, the AUC of ROC increased from $0.76$ (only eviction data) to $0.88$ (all features).

Similar to the Pre-COVID models, we observe a higher gain for the Post-COVID (XGB) model after the inclusion of the neighborhood attributes ($0.73$ to $0.84$) and an additional gain of $5$ points after including the owner attributes ($0.84$ to $0.90$) (Figure~\ref{fig:xgb_roc_21}) for the ROC AUC for $T_{3}$. We note that all the AUC differences in Pre-COVID and Post-COVID models are statistically significant at $p< 0.001$.

\textbf{Performance Remains Stable for Different Models.}
We compare three predictive models to predict evictions where two of them are based on decision trees (Random Forest and XGBoost) and one of them is a Feedforward Neural Network. The performance remains stable for all of the models for most of the test periods. Although we observe differences in a few cases, for example, the difference in AUC of ROC between Pre-COVID (XGB) and Pre-COVID (FNN) for predicting $T_{2}$ using eviction and neighborhood features is $8$ points, the difference is negligible when all of the features are included in the prediction where the AUC of ROC remains between $0.88$ to $0.92$ (three right most columns in the table). Most of the AUCs of PR also remain within $0.42$ and $0.45$ when using all features. 

\textbf{Models Perform Better in the Distant Future When Owner and Neighborhood Features are Used.}
We now turn to understanding how well models generalize over time, and we do this in two quite different contexts. In general, we would expect performance to degrade over time (that is, a model trained on data from January - October 2021 would, all else equal, be expected to perform better in predicting evictions from November 2021 - January 2022 than in predicting evictions from November 2022 - January 2023). However, with the dramatic impact of the eviction moratorium caused by COVID-19 on eviction patterns, we would expect to see a more substantial degradation in performance if the two test periods are on different sides of the eviction moratorium. 

Our results in Table~\ref{tab:comparative_results} demonstrate this clearly when only the eviction features are used. The performance of Pre-COVID model (XGB) drops from an AUC ROC of $0.76$ to $0.71$ and an AUC PR of $0.38$ to $0.36$ when predicting $T_2$ rather than $T_1$. The drop is not observed when using a similar ``delay'' but entirely in the Post-COVID timeframe. Interestingly, we do not see a substantial drop in performance when using owner and neighborhood attributes with the eviction records, showing that these attributes add robustness to the models, even during an unprecedented emergency like COVID-19, compared with simply using eviction records as features.

%% file: tables/table-2.tex
\begin{table*}
% \normal
\centering
\def\arraystretch{1.4}%
\caption{Cross-model performance for predicting tenant evictions.  \textit{Training Period} indicates whether modeling occurs before or after the pandemic onset. \textit{Testing Period} and \textit{Base Rate} show the eviction rate during predictions. \textit{Feature Sets} report AUC from Random Forest (RF), XGBoost (XGB), and Feedforward Neural Network (FNN) classifiers when including additional information across testing periods. The top rows summarize AUC for the ROC curves, while precision-recall curves are presented at the bottom. We note that AUC improves significantly ($p < 0.001$) with each additional feature set for most testing periods.}

  \begin{tabular}{|p{2.7 cm}|c|p{0.8 cm}|c|c|c|c|c|c|c|c|c|}
 
   \hline 
    \centering
    \multirow{3}{*}{Train Period} &\multirow{3}{*}{\thead[c]{Test \\ Period}} &\multirow{3}{*}{\thead[c]{Base \\Rate}} &
      \multicolumn{9}{c|}{Feature Sets}\\
    \cline{4-12}
    & & &\multicolumn{3}{c|}{\thead[c]{Eviction}}&
    \multicolumn{3}{c|}{\thead[c]{Eviction and \\Neighborhood}}
     &\multicolumn{3}{c|}{\thead[c]{Eviction, Neighborhood, \\and Owner}}\\
      \cline{4-12}
      % \hline
     & & &RF &XGB &FNN &RF &XGB &FNN &RF &XGB &FNN\\
     \hline
     \multicolumn{12}{|c|}{AUC for ROC Curve}\\
    \hline 
    % \midrule  
    \multirow{3}{*}{\thead[c]{Pre-COVID\\Jan 2019  - Oct 2019}} & $T_{1}$ & 0.032&0.76 & \textbf{0.76} &0.76 & 0.82 & \textbf{0.84} &0.82 & 0.88 &\textbf{0.89}& 0.88\\
    \cline{2-12}

     & $T_{2}$ & 0.019 &  0.71 &\textbf{0.71}&0.71& 0.80 &0.86&0.78& 0.90& 0.92&0.89\\
 
    \cline{2-12}

     & $T_{3}$   & 0.021 & 0.73 &0.73 &0.73& 0.83 &0.84 &0.82& 0.89 &0.90&0.89\\

    \hline  
    \multirow{2}{*}{\thead[c]{Post-COVID\\ Jan 2021 - Oct 2021}}& $T_{3}$ &0.021 & 0.75 & \textbf{0.73} & 0.74& 0.83 &\textbf{0.84}& 0.81 & 0.89&\textbf{0.90} &0.90\\
    % \hline
    \cline{2-12}
     & $T_{4}$  &0.019 & 0.75 & 0.75 &0.75& 0.83 & 0.84 & 0.80& 0.89&\textbf{0.89} & 0.89 \\
  
    \hline
    \multicolumn{12}{|c|}{AUC for Precision Recall Curve}\\
    \hline
        \multirow{3}{*}{\thead[c]{Pre-COVID\\Jan 2019 - Oct 2019}} & $T_{1}$ & 0.032 & 0.37 &\textbf{0.38}& 0.38 & 0.42&\textbf{0.43}& 0.42 & 0.45 &\textbf{0.47} & 0.45\\
    \cline{2-12}

     & $T_{2}$ &0.019 & 0.37 &\textbf{0.36} & 0.32 & 0.39 &0.37&0.33 & 0.40 & 0.44 &0.42\\
 
    \cline{2-12}

     & $T_{3}$  & 0.021 & 0.39 &0.39& 0.33 & 0.36&0.39& 0.36 &0.42 & 0.43 &0.41\\

    \hline
    \multirow{2}{*}{\thead[c]{Post-COVID\\Jan 2021 - Oct 2021}}& $T_{3}$  & 0.021 & 0.39 & 0.38 & 0.32 & 0.36 & 0.36  & 0.34 &0.40 & 0.40 &0.40\\
    % \hline
    \cline{2-12}
     & $T_{4}$  &0.019 & 0.39 &0.38& 0.33 & 0.37 & 0.37& 0.36 & 0.40 &\textbf{0.42} & 0.41\\
    \hline
    
    \multicolumn{12}{l}{%
    \begin{minipage}{14.5cm}%
   \small{$^*$ RF = Random Forest Classifier, XGB = XGBoost Classifier, FNN = Feedforward Neural Network}
     \end{minipage}%
}\\
   \end{tabular}
   \label{tab:comparative_results}
\end{table*}

%% file: sections/4-Outreach.tex
\section{Outreach Routing Policy}
We now turn back to one of the main questions of interest: Can we leverage the generated property-level risk scores from the models to devise better routing policies for caseworkers to visit households that may be at high risk of eviction? 

We describe our risk-score-based outreach policy and compare its performance in discovering evictions with commonly used policies. 

\subsection{Routing Path and Estimated Outreach Time}
To compare outreach policies, we require an estimate of the time required to canvass a set of properties. We estimate the time it takes to canvass property $B$ from property $A$ in two parts: first, we estimate the time it takes to travel from $A$ to $B$ using a piecewise linear function; second, we add a cost proportional to the unit count of $B$ to account for knocking on all doors upon arrival. This cost is calibrated to data provided by community partners. Further details of the implementation are provided in \Cref{sec:routing_details}. The geodesic distance between two properties is calculated using the GeoPy Python library \cite{geopy}, and the fastest route between all canvassed properties is calculated as an approximate solution to the Traveling Salesperson Problem (TSP) using the OR-tools library \cite{ortools}.

\subsection{Routing Policies}

\subsubsection{Risk Score (RS) Based Outreach}
We consider risk scores generated by XGBoost -- the best-performing model -- for $T_4$. This presented the latest period (November 2022 to January 2023) available in the dataset and is most interesting to our community partners.  

\begin{figure}[h]
    \centering 
    \includegraphics[width=0.85\columnwidth]{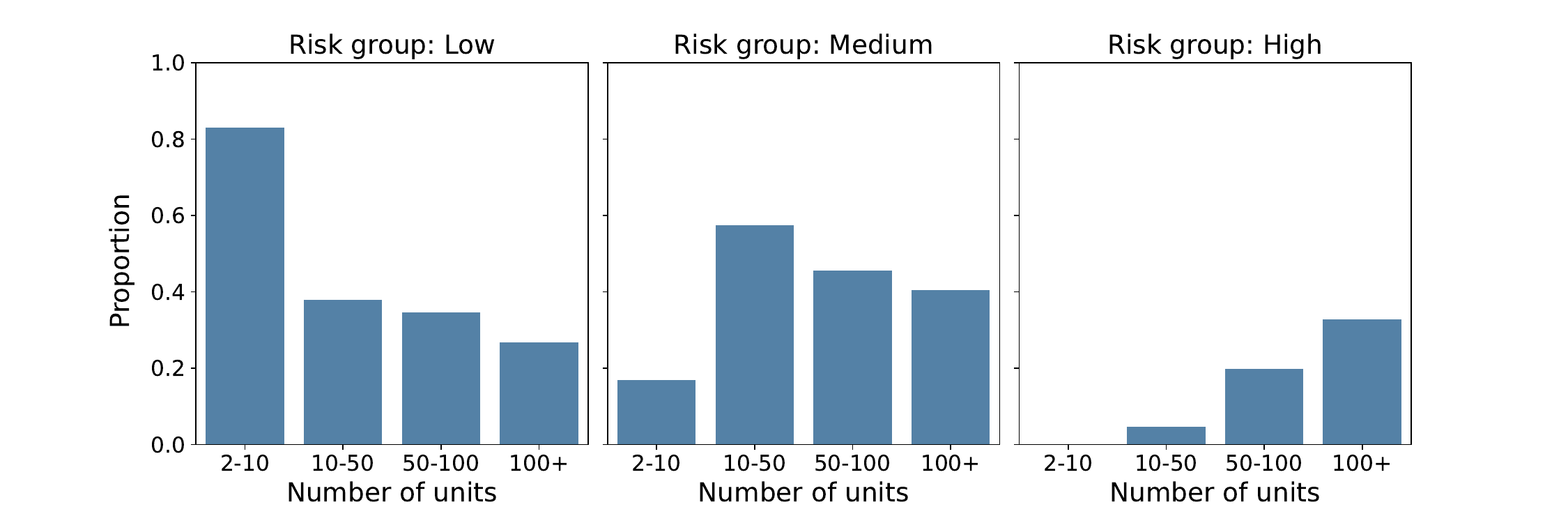}
    \caption{Distribution of number of units that fall in a particular risk group [Low, Medium, High]. The X-axis represents the groups based on the number of units, and the Y-axis represents the proportion of properties from each category of unit sizes.}  
    \label{fig:unit_number_vs_risk_groups}
\end{figure}

We categorize the risk scores into four groups (Very Low, Low, Medium, and High) for identifying properties with a greater risk of eviction. Figure~\ref{fig:unit_number_vs_risk_groups} shows the distribution of properties in each risk group based on the number of units in the property. Most properties in $T_{4}$ ($69.3\%$) fall into the Very Low category, where risk scores range from $0$ to $0.05$, reflecting a true eviction rate of $0.003$. The ranges of risk scores for the Low, Medium, and High-risk groups are $(0.05-0.2],(0.2-0.8]$ and $(0.8-1.0]$. The percentages of properties in these groups and corresponding true eviction rates are $(22.08\%,7.59\%,0.98\%)$ and $(0.01,0.09,0.61)$, respectively. We might worry that the model is simply picking up on the existence of higher numbers of units in a property to assign higher risk scores, but, even within a unit size category (e.g., 100+ units), we see significant variance among the risk score groups assigned to different properties, which alleviates this concern (see~\Cref{fig:unit_number_vs_risk_groups}). 

\textbf{Neighborhood, Eviction, and Owner (based) - Targeted Outreach (NEO-T-O)}. We select properties in the medium and high-risk groups for targeting outreach efforts (i.e., caseworkers). There are 2,299 properties situated in St. Louis City and County in these two categories. To approximate the amount of time it would take to canvass each unit in these properties, as described above, we obtain an approximate solution of the TSP to estimate travel time, and add to that the estimated time spent ``knocking on doors''.

\textbf{RS Outreach Based on Eviction Records Only.}
We consider risk scores generated by XGB using features only from historical eviction records. We sort the properties based on these risk scores and select the top-$k$ properties that can be visited in the same time as that spent by NEO-T-O in order to ensure fair comparison ($k$ is estimated using incremental search).

\textbf{RS Outreach Based on Eviction and Neighborhood Information.} This policy is the same as the one above, except that it considers risk generated by XGB using features from both eviction records and neighborhood information.

\subsubsection{Alternative Policies}
We compare NEO-T-O and the other two RS outreach policies with two common outreach practices: 
targeting based on prior evictions (regardless of property location), and canvassing all properties within specific neighborhoods. 
Again, in order to ensure a fair comparison, we control for available outreach time. These policies end up looking quite different from the ones above, in that they tend to visit many more (smaller) properties, rather than concentrating resources on fewer larger properties. Thus, they spend more of their time traveling between properties and are able to visit fewer overall units. In order to separately analyze the effect of this substantive difference, we also consider variants that visit the same number of \emph{units} as NEO-T-O.

\textbf{Previous Eviction Count Based Policy.}
The properties are sorted by the number of prior evictions in a previous period, and outreach targets the ``top-$k$'' properties in terms of prior evictions, again computing travel time by approximating the TSP solution among these $k$. We use observed filings in the quarter (August 2022 - October 2022) prior to period $T_{4}$.

\textbf{Neighborhood Based Policy.} This routing policy visits all rental properties in certain neighborhoods. Neighborhoods -- defined by census block groups -- are ranked by prior eviction counts observed in the quarter before $T_{4}$. For each neighborhood, we compute the approximately optimal route to visit all rental properties. We continue visiting neighborhoods in descending order until the time or unit constraints are met.

\subsection{Outreach Routing Results and Analysis}\label{sec:outreach_results}

\textbf{Usefulness of Risk Scores.}
The headline result is that the total number of evictions discovered (936) by NEO-T-O is $8.5\%$ higher than the neighborhood-based policy (863) and $28\%$ higher than the previous eviction-based policy (731) when controlling for outreach time. 
While risk-score-based policies discover more evictions than alternative policies, they canvass a substantially lower number of properties despite both types of policies visiting similar numbers of units. Thus, RS-based policies are more useful if outreach programs want to utilize limited resources (e.g., caseworkers) to discover more evictions while canvassing a small number of properties. Both policies canvass a similar number of units, which implies that RS-based policies target large properties with a higher number of units, whereas alternative policies also include properties with fewer units.
For the same reason, RS-based policies yield higher eviction discovery rates on a per-property basis than alternative policies.

\input{tables/table-3}
\begin{figure}
    \centering
    \captionsetup[subfigure]{justification=centering}
    \subfloat[Properties visited based on NEO-T-O (blue), previous eviction-based policy (orange), and both policies (red).]{\includegraphics[width=0.3\columnwidth]{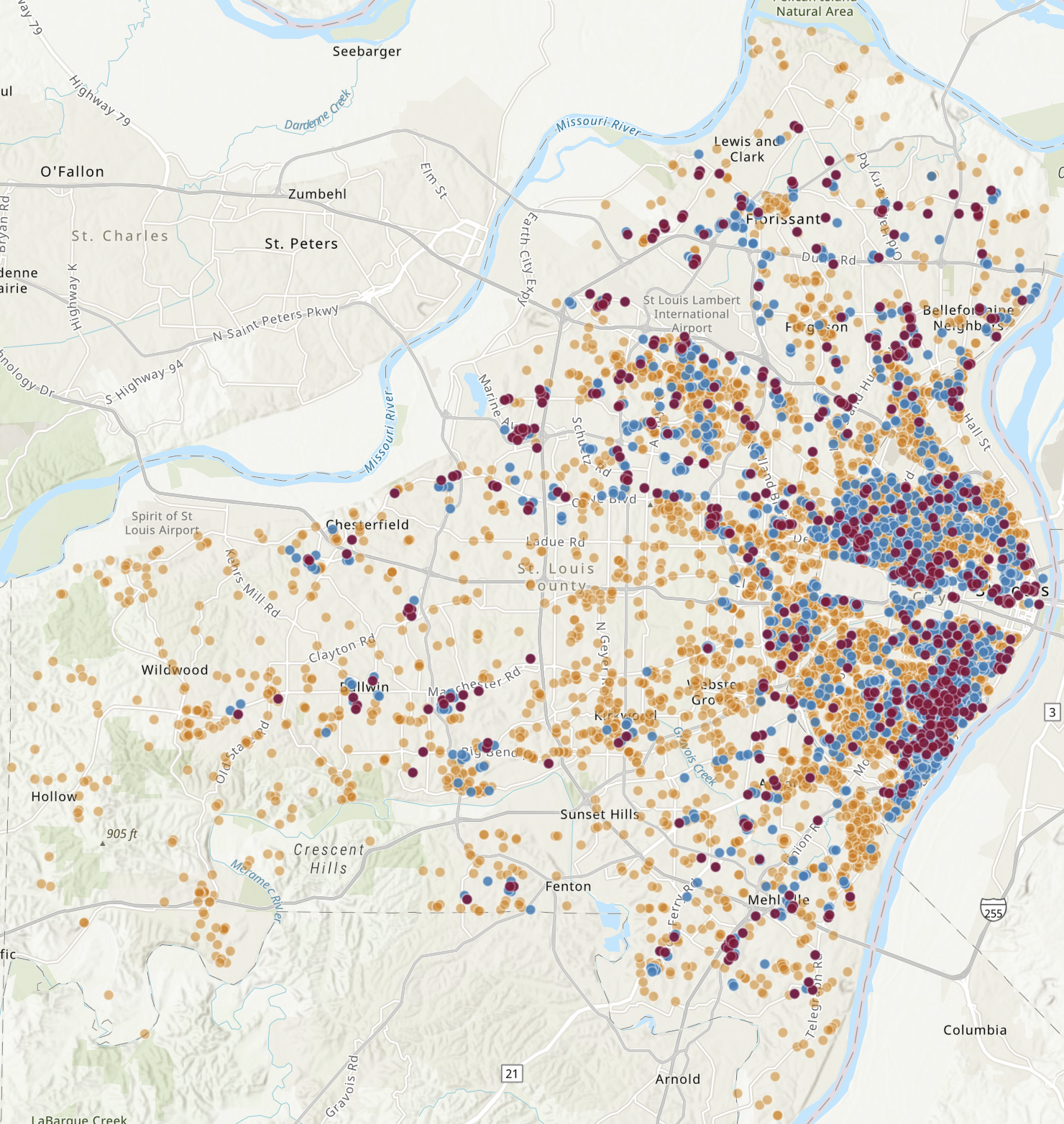}\label{fig:panel_a}}
    \hspace{3mm}
    \subfloat[Properties visited based on NEO-T-O (blue), neighborhood-based policy (orange), and both policies (red).]{\includegraphics[width=0.27\columnwidth]{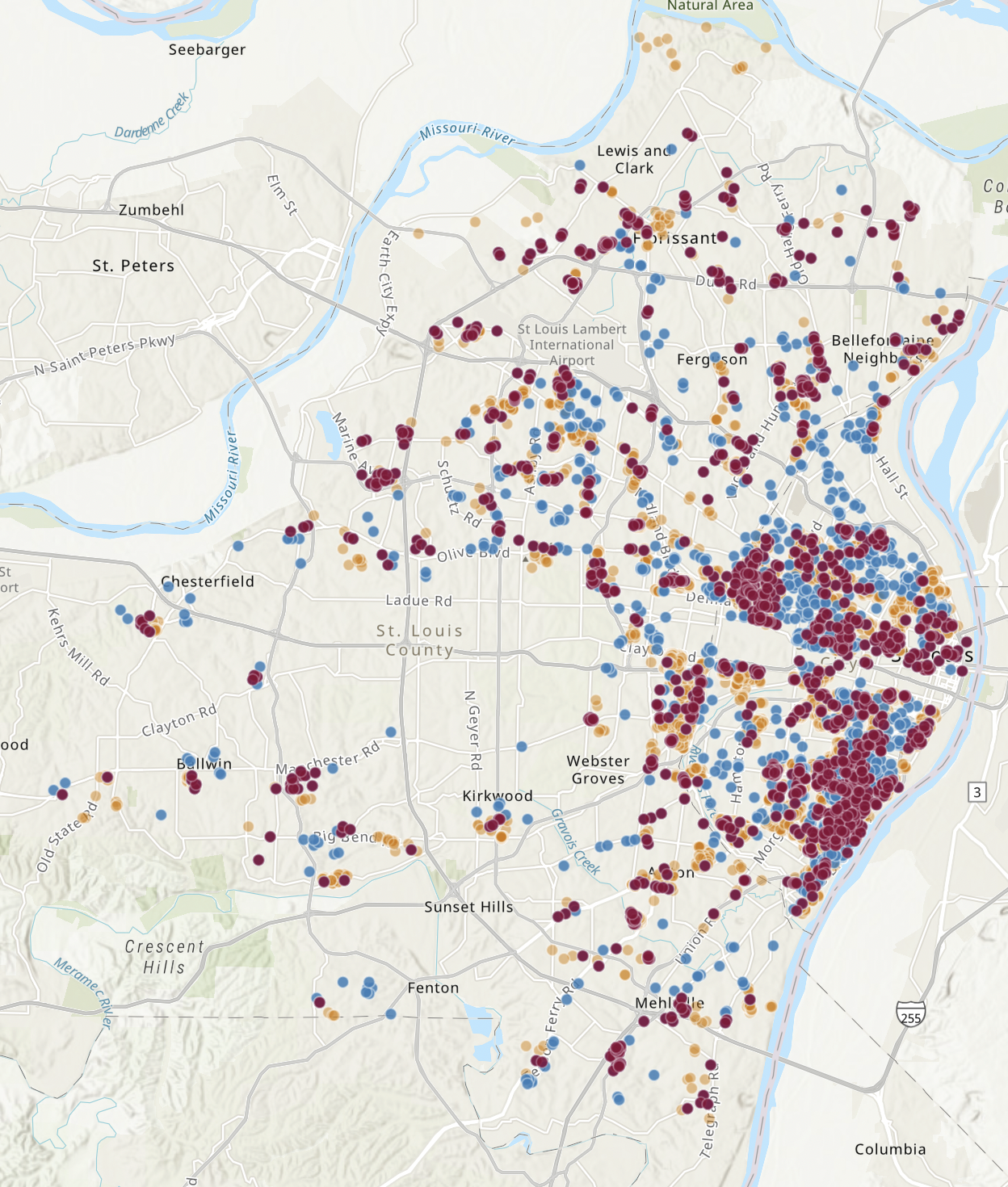}\label{fig:panel_b}}   
    \hspace{5mm}
    \subfloat[Properties visited by NEO-T-O policy, neighborhood-based policy and both in St Louis City.]{\includegraphics[width=0.3\textwidth]{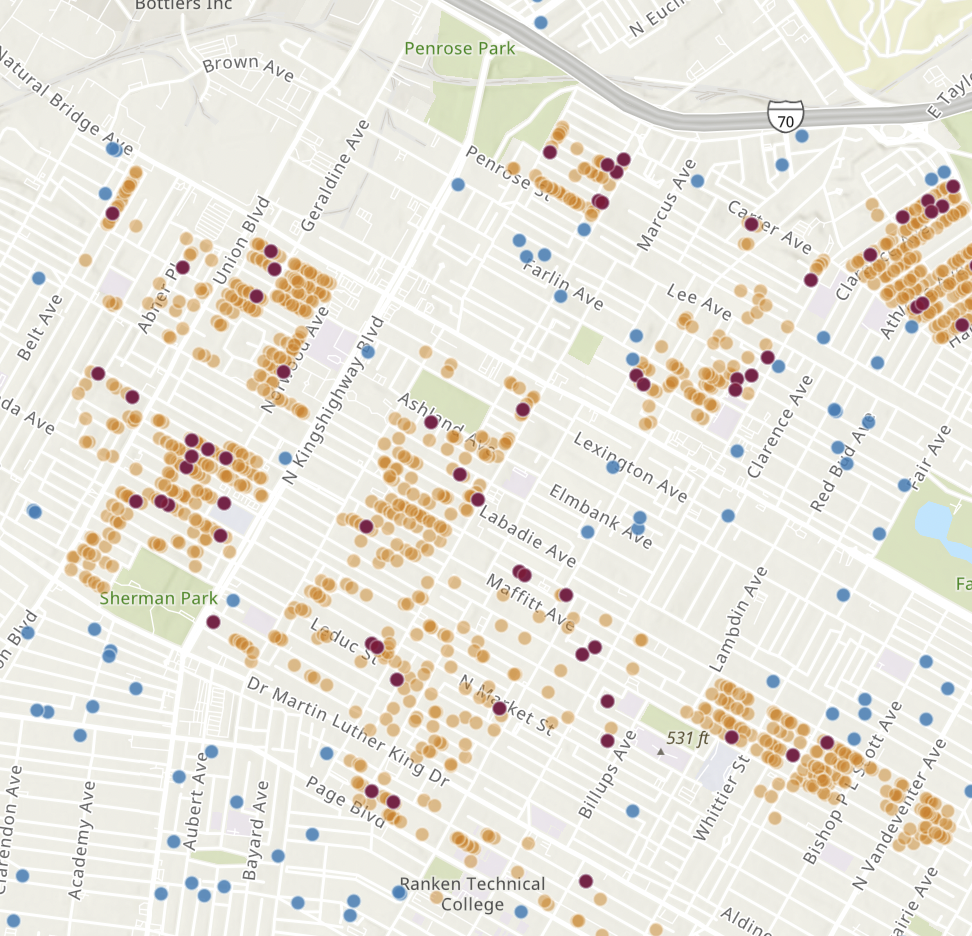}\label{fig:panel_c}}
    \caption{Canvassed properties in St Louis City and County using different outreach policies. In all panels, blue dots represent the properties canvassed by NEO-T-O. Panel (a) shows properties selected by previous eviction-based policy (orange dots) - scattered across the map - and properties canvassed by both NEO-T-O and previous eviction-based policy (red dots). Panel (b) shows the properties selected by neighborhood-based policy (orange dots) - grouped in different areas - and properties canvassed by both NEO-T-O and neighborhood-based policy (red dots). Panel (c) shows a close-up of part of Panel (b).}
 
    \label{fig:map}
\end{figure}

\textbf{Importance of Neighborhood and Owner Features.} The addition of neighborhood features yields substantial marginal value to the risk scores when used in our routing policy, while the further addition of owner attributes has limited additional value beyond that. When adding neighborhood features with the eviction features, the number of discovered evictions increases to $919$ from $677$, yielding a $35.8\%$ gain. However, adding owner features to the eviction and neighborhood features yields only a $1.8\%$ gain (from 919 to 936). It is also worth noting here that the marginal benefit to the outreach strategy is better captured by the area under the ROC curve as opposed to the area under the Precision-Recall curve. This is contrary to the general conjecture in the literature that the PR curve has more value in discovery tasks with low base rates \cite{davis2006relationship,sofaer2019area}.

\textbf{Visualizing the Policies.}
Figure~\ref{fig:map} 
% {\color{red} figure captions remain confusing}
shows maps of St. Louis City and County, where the dots indicate properties that would be reached by different outreach policies. We plot the properties canvassed by NEO-T-O and two categories of alternative policies when controlling for time: previous eviction-based outreach and neighborhood-based outreach.

We observe that the 13,122 properties canvased through the eviction-based policy are scattered across the city (\Cref{fig:panel_a}), whereas the 11,460 properties of the neighborhood-based policy are (by design) grouped in several census blocks of the map (\Cref{fig:panel_b}). 582 properties are canvassed by both NEO-T-O and the prior eviction-based policy, covering $25\%$ of all properties canvassed by NEO-T-O. 1,419 properties are canvassed by both NEO-T-O and the neighborhood-based policy, covering $62\%$ of the NEO-T-O properties. This large intersection between the properties targeted by NEO-T-O and the neighborhood-based policy further explains why we do not see a large increase in the number of evictions discovered by NEO-T-O in comparison to the RS-based policy using previous eviction and neighborhood information. \Cref{fig:panel_c} shows a closer look at the properties reached by NEO-T-O (blue), neighborhood-based policy (orange), and both (red) policies. While the orange dots cover all the properties within a neighborhood, we observe some properties near the boundary of the neighborhoods that are not canvassed by the neighborhood-based policy but have high risks of eviction (blue) according to NEO-T-O.

%% file: tables/table-3.tex
\begin{table*}
\centering
% \normal
\def\arraystretch{1.4}%
\caption{
Cross-policy performance analysis for discovering tenant evictions. \textit{Routing Policy} lists the risk-score-based and alternative outreach policies we consider. 
\textit{Control} indicates the controlling factors in policies.
\textit{Normalized Outreach Time} shows the outreach time normalized by the time required by NEO-T-O. \textit{Eviction Discovery
Rate} reports the rate of eviction discovery per property for each policy. \textit{Units Visited} reports the total number of units that are visited by the policy.}

\begin{center}
  \begin{tabular}{|P{4.1 cm}|P{1.35cm}|P{1.35cm}|P{1.5cm}|P{1.2cm}|P{1cm}|P{1.45 cm}|}
  \hline
  \centering

  Routing Policy & Control  
  & Normalized Outreach Time &  Evictions Discovered
  &  Properties Visited & Units Visited  & Eviction Discovery Rate\\
\hline
\multicolumn{7}{|c|}{  
  Risk Score Based Policy}\\
\hline

NEO-T-O & Time
& 1
& 936
& 2,299
& 101,119
& 40.7\%
\\
\cline{1-7}

Eviction Features & Time
& 1
& 677
& 1,711
& 101,288
& 39.6\%
\\
\cline{1-7}

Eviction and Neighborhood Features & Time
& 1
& 919
& 4,000
& 94,556
& 23.0\%
\\
\hline

\multicolumn{7}{|c|}{  
  Alternative Policies}

\\ 
\cline{1-7}
\multirow{2}{*}{\thead[c]{Previous Eviction Count}}
& Time
& 1
& 731
& 13,122
& 90,943
& 5.6\%\\
\cline{2-7}

& Unit
& 1.63
& 749
& 14,987
& 101,118
& 5.0\%
\\
% \hline

\cline{1-7}
\multirow{2}{*}{\thead[c]{Neighborhood Based}}
& Time
& 1
&863
&11,460
& 99,532
& 7.5\%
\\

\cline{2-7}
& Unit
& 1.03
& 870
& 11,905
& 101,116
& 7.3\%
\\
\hline

   \end{tabular}
   \label{tab:comparative_results_strategy}
\end{center}

\end{table*}

%% file: sections/5-Discussion.tex
\section{Discussion}
What are the benefits of using risk scores in societal service provisions such as the prevention of eviction? How should we evaluate these risk scores in real-world decision-making? This paper adds to the growing literature \cite{Wilder_Dilkina_Tambe_2019} making the case that we need to go beyond prediction accuracy and critically examine how usable the risk scores are in \emph{decision making} -- in our case, devising outreach programs and preventing evictions. 

In some settings, individualized risk scoring, even when relatively accurate, has not been shown to have benefits in terms of interventions beyond those that could be obtained through less granular means \cite{perdomo2023difficult}. We obtain a somewhat different result in a different domain. 
We propose a risk-score-based policy -- NEO-T-O -- that identifies eviction-prone properties to send a theoretical set of caseworkers to discover possible evictions and provide assistance. Our results show that such a policy is, in fact, more useful in discovering evictions than common outreach practices that focus exclusively on certain neighborhoods or properties with recent prior eviction histories. 
However, it is also worth noting that, consistent with existing research, a significant portion of the benefit can be achieved through the use of neighborhood socio-demographic information. 

Future research should be aware of trade-offs in using risk-score-based outreach policies. NEO-T-O focuses on a small number of properties with high eviction risk. This means that NEO-T-O coverage is focused on developments with large numbers of units, which may be efficient in the short term but leave out certain populations or lead to systemic problems downstream.
%smaller than alternative policies such as neighborhood-based (NB) outreach, which covers all of the properties in the area during canvassing. It is possible that NEO-T-O can ignore some properties in an area while over-surveilling the targeted properties, which could result in unfair canvassing. 
There is also a theoretical risk that risk-score-based policies could be prone to feedback loops~\cite{ensign2018runaway}: caseworkers are sent to targeted properties, leading to higher rates of discovery of potential evictions, and subsequently, more caseworkers are sent to those properties, and so on.

Our results also show that risk scores that are critical for NEO-T-O are typically stable across different models and timeframes (including pre-and post-COVID timeframes). Classifiers accurately identify properties at risk of eviction up to 24 months before filing when incorporating publicly available information on past evictions, neighborhood characteristics, and owner features. The evidence suggests a relatively wide window of opportunity for tenant-landlord intervention. Moreover, the unique data linkages identify property ownership registered under various names and corporate entities across the region. The information could enable greater scrutiny of nefarious owner eviction behaviors, housing code violations, tax abatement, and other public funding that could inform strategic action by governmental and non-governmental agencies to protect tenants from problematic owners~\cite{seymourBuildingEvictionEconomy2021,garbodenSerialFilingHow2019,greifRegulatingLandlordsUnintended2018}.

Our findings must be interpreted under several limitations. The results are currently limited to one city and county. This is primarily because curating such a dataset is itself a challenging task due to the presence of restrictive state and local policies and limitations of governmental infrastructures. We found that eviction itself is a relatively low-probability event and generally occurs in rental properties owned by larger corporations rather than single-owner family homes. Thus, our analysis is limited to residential rental properties with two or more units. In our experiments, the data only represent legal eviction filings; illegal evictions in which owners displace tenants without court orders remain unobserved.  
% Advocates report rampant illegal lockouts that require further investigation.
We also note that we do not have any personal information about individual tenants in the data. The eviction court orders do not contain sensitive tenant information, such as gender, race, etc., other than their names. Thus, it is difficult to evaluate the responses of the routing policies towards different social groups.

%% file: sections/6-Statements.tex
\newpage

\paragraph{\textbf{Research Ethics Statement}} The research accords with the ACM Code of Ethics and Professional Conduct. We intentionally engage community partners in framing the problem, and the research team appropriately includes individuals with appropriate knowledge and skills for implementing the study design. Although the data come from publicly available information sources, we use a data-sharing agreement between community partners to ensure appropriate use. Our modeling generates property-level eviction risk scores; however, these are not currently shared with community partners, and evaluation relies on simulation to ensure privacy and avoid unintended consequences from deployment without adequate testing. Subsequent research will examine how to implement methods in ways that improve well-being, avoid harm, and demonstrate trustworthiness. 

% Predicting eviction risks can have a significant impact on directing appropriate assistance to tenant households that are at higher risk of eviction. In our study, we show the potential effect of
% ownership in predicting eviction risk scores that result in accuracy improvements. However, the study was limited to multi-unit properties, not including the single units where evictions might be less frequent and smaller property owners might face more foreclosures and property sales that result in evictions of their current tenants.

% Risk focusing on improvements lead societal push toward incremental

\paragraph{\textbf{Researcher Positionality Statement}}
% \textcolor{blue}{@Patrick can you take a look at what we can write here reflections on how our background and experiences inform or shape the work. -- Tasfia} 
 The community-driven research is part of an ongoing partnership with local stakeholders working to prevent evictions and the deleterious impact on children, families, and communities. The authors include a diverse trans-disciplinary team (computer science, data science, community public health) that includes doctoral students and faculty from around the globe. Members approach the work from different perspectives. Active engagement of community partners through frequent meetings (multiple times per week when necessary) provides a strong foundation for repeatedly assessing the utility, feasibility, and ethical implications of the data and modeling. Our partners include another diverse set of experts with extensive experience in service provision, policy, and navigating unstable housing in tight affordable housing markets. The work aims to acknowledge and leverage the various perspectives into building technical and equitable solutions for keeping low-income families housed. 

\paragraph{\textbf{Adverse Impact Statement}} 
Our study demonstrates the feasibility and potential utility of risk-score-based outreach to prevent tenant evictions. Simulation suggests AI-enabled tools could make targeting more efficient, yet deployment requires careful consideration of unintended consequences. The scarcity of housing resources forces difficult decisions regarding who to assist that must account for equity across individuals and groups. Moreover, outreach efforts that alienate rental owners could further deplete access to affordable housing. 
The implementation of risk-score-based outreach that prevents eviction must be carefully and collaboratively designed and evaluated -- a goal of our future research.

%% file: sections/7-Appendix.tex
\appendix

\section{Appendix A}
\label{sec:appendix_a}
\subsection{Model Architecture and Hyperparameters}
\label{sec:model_details}
We compare performance across three classifiers -- Random Forest (RF), XGBoost (XGB), and Feedforward Neural Network (FNN) -- with hyperparameters set via grid search to maximize precision. 
\subsubsection{Hyper-parameters for Eviction Prediction Using Random Forest} 

For Random Forest, we performed a grid search based on the following hyper-parameters:
\begin{itemize}
    \item Measure of impurity: Gini, Entropy, Log Loss 
    \item Max depth: [2, 5, 10, 15]
    \item Minimum samples in a leaf node: [1, 2, 3, 4, 5]
    \item n\_estimators: [50, 100, 500]
\end{itemize}

\begin{table*}[!htp]
\caption{The combination of hyper-parameters for Random Forest Models}
\centering
\def\arraystretch{1.4}%
\begin{tabular}{|c|P{4cm}|P{1.8cm}|P{1.8cm}|P{1.8cm}|P{1.8cm}|}
\hline
 \multirow{2}{*}{Model} & \multirow{2}{*}{Features}& \multicolumn{4}{c|}{Parameters}\\
 \cline{3-6}
 & & Measure of impurity & Max depth & Minimum samples in a leaf node & n\_estimators\\
 \hline
 \multirow{3}{*}{Pre-COVID} & Eviction &Entropy &5 &2 &500 \\
 \cline{2-6}
 & Eviction and Neighborhood & Log Loss & 5 & 2 & 500 \\
 \cline{2-6}
 & Eviction, Neighborhood, and Owner & Entropy & 5 & 2 & 500 \\
\hline

\multirow{3}{*}{Post-COVID} & Eviction & Gini & 3 & 2 & 100 \\
 \cline{2-6}
 & Eviction and Neighborhood &Log Loss & 2 &2 & 100\\
 \cline{2-6}
 & Eviction, Neighborhood, and Owner &log Loss & 5 & 2& 100 \\
 \hline
\end{tabular}
\end{table*}

\subsubsection{Hyper-parameters for Eviction Prediction Using XGBoost}
For XGBoost, we performed a grid search based on the following hyper-parameters:
\begin{itemize}
    \item Max depth: [2, 3, 4, 5]
    \item Learning Rate: [0.01, 0.05, 0.1]
    \item n\_estimators: [50, 100, 500]
    \item scale\_pos\_weight: [1, 3, 5]
    \item gamma: [0, 0.05, 0.1]
\end{itemize}

\begin{table*}[!htp]
\caption{The combination of hyper-parameters for XGBoost Models}
\centering
\def\arraystretch{1.4}%
\begin{tabular}{|c|P{4 cm}|P{1.3cm}|P{1.3cm}|P{1.6cm}|P{1.5cm}|P{1.3cm}|}
\hline
 \multirow{2}{*}{Model} & \multirow{2}{*}{Features}& \multicolumn{5}{c|}{Parameters}\\
 \cline{3-7}
 & & Max depth & Learning Rate& n estimators &scale pos weight & gamma \\
 \hline
 \multirow{3}{*}{Pre-COVID} & Eviction & 3 & 0.05 & 100 & 1&0.1 \\
 \cline{2-7}
 & Eviction and Neighborhood &3 & 0.1 & 100 &1 & 0.1\\
 \cline{2-7}
 & Eviction, Neighborhood, and Owner &2 & 0.1&100& 1 &0.1 \\
\hline

\multirow{3}{*}{Post-COVID} & Eviction &3 & 0.05 & 100 & 3&0.05\\
 \cline{2-7}
 & Eviction and Neighborhood &3 & 0.05& 100 &1 &0.05\\
 \cline{2-7}
 & Eviction, Neighborhood, and Owner & 3 & 0.05& 100 & 5&0.05\\
 \hline
\end{tabular}
\end{table*}

\subsubsection{FNN Architecture}
The FNN architecture consists of an input layer, two hidden layers, and an output layer. The input layer is designed with several neurons equal to the input features, while the first hidden layer comprises 64 neurons activated by rectified linear units (ReLU), followed by Batch Normalization for normalization and Dropout with a rate of 0.5 for regularization. The second hidden layer conducts the same processing with 32 neurons. %The second hidden layer consists of 32 neurons with ReLU activation, Batch Normalization, and Dropout with a similar rate. 
The output layer uses a single neuron with the sigmoid activation function. This architecture is compiled using binary cross-entropy as the loss function and the Adam optimizer.

\section{Appendix B}
\label{sec:appendix_b}
\subsection{Routing Path and Estimated Outreach Time}
\label{sec:routing_details}

We use geodesic distance from the GeoPy Python library~\cite{geopy} to estimate the distance between properties from the latitudes and longitudes of the properties. We also evaluated OpenStreetMap~\cite{OpenStreetMap} for calculating the distances. The Pearson correlation between the distances calculated using both methods on a random sample (100,000 pairs of properties) was 0.98. We decided to use geodesic distance as it is significantly faster than OpenStreetMap, and we have a relatively large number of property pairs for potential targeting (see~\autoref{sec:outreach_results}).
With the distances, we used the OR-tools library by Google~\cite{ortools} to obtain an approximately optimal route to visit a set of properties. We record the estimated time (i.e., outreach time) to complete visits to the properties.
We consider the following piece-wise speed function to estimate the time required to travel from one property to another. The speed gradually increases as we increase the distance. Miles (m) represents the unit for the distance and mile-per-hour (mph) for speed. For distance $d$, speed $s$:

\begin{align}\label{eq:speed_func}
    s (\text{mph}) = \begin{cases} 
      4 & d \leq 1 \\
      15 & 1< d\leq 3 \\
      % 25 & 10< d\leq 15 \\
      30 & 3< d\leq 5 \\
      55 & d > 5     
   \end{cases}
\end{align}

We estimate the actual time taken during a visit to a unit as $n$ hours. We set $n$ to $6$ minutes or $0.1$ hours. The estimate was based on data collected by community partners during door-to-door canvassing to inform tenants of COVID-19 rental relief. Canvassers knocked on an average of 35 doors during six-hour shifts, speaking with residents and/or leaving flyers with resource information. For a pair of properties $p_{1}, p_{2}$ with units $u_{1}, u_{2}$, and distance $d_{1,2}$, the total time $t(p_1,p_2)$ to visit properties $p_{1}$ and $p_{2}$ is, 
\begin{align}\label{eq:dist_func}
    t(p_{1},p_{2}) = n * u_{1} + s * d_{1,2} + n * u_{2}
\end{align}

Thus, the routing policies vary by the time needed to travel between properties. 
%Door-to-door canvassing requires less travel time given the sequential nature, whereas targeted outreach sends canvassers to the closest and highest-risk property that may not be next door or on the same street. 